\documentclass[journal]{vgtc}                     % final (journal style)

\usepackage{listings}

\onlineid{0}

\vgtccategory{Research}

\vgtcpapertype{Representation and Interaction}

\title{Can LLMs Generate Visualizations with Dataless Prompts?}

\author{Darius Coelho, Harshit Barot, Naitik Rathod, Klaus Mueller
}

\authorfooter{
 \item Darius Coelho, Harshit Barot, Naitik Rathod, and Klaus Mueller are with Stony Brook University.
E-mail: \{dcoelho | hbarot | nrathod | mueller\}@cs.stonybrook.edu
}

\abstract{%
Recent advancements in large language models have revolutionized information access, as these models harness data available on the web to address complex queries, becoming the preferred information source for many users.
In certain cases, queries are about publicly available data, which can be effectively answered with data visualizations.
In this paper, we investigate the ability of large language models to provide accurate data and relevant visualizations in response to such queries. 
Specifically, we investigate the ability of GPT-3 and GPT-4 to generate visualizations with dataless prompts, where no data accompanies the query.  
We evaluate the results of the models by comparing them to visualization \textit{cheat sheets} created by visualization experts. 
}

\keywords{Large Language Models, Data Visualization, Captions, Prompt Engineering}

\teaser{
  \centering
  \includegraphics[width=\linewidth, alt={A visualization chart sheet to show which chart can be generated using the type of data in consideration}]{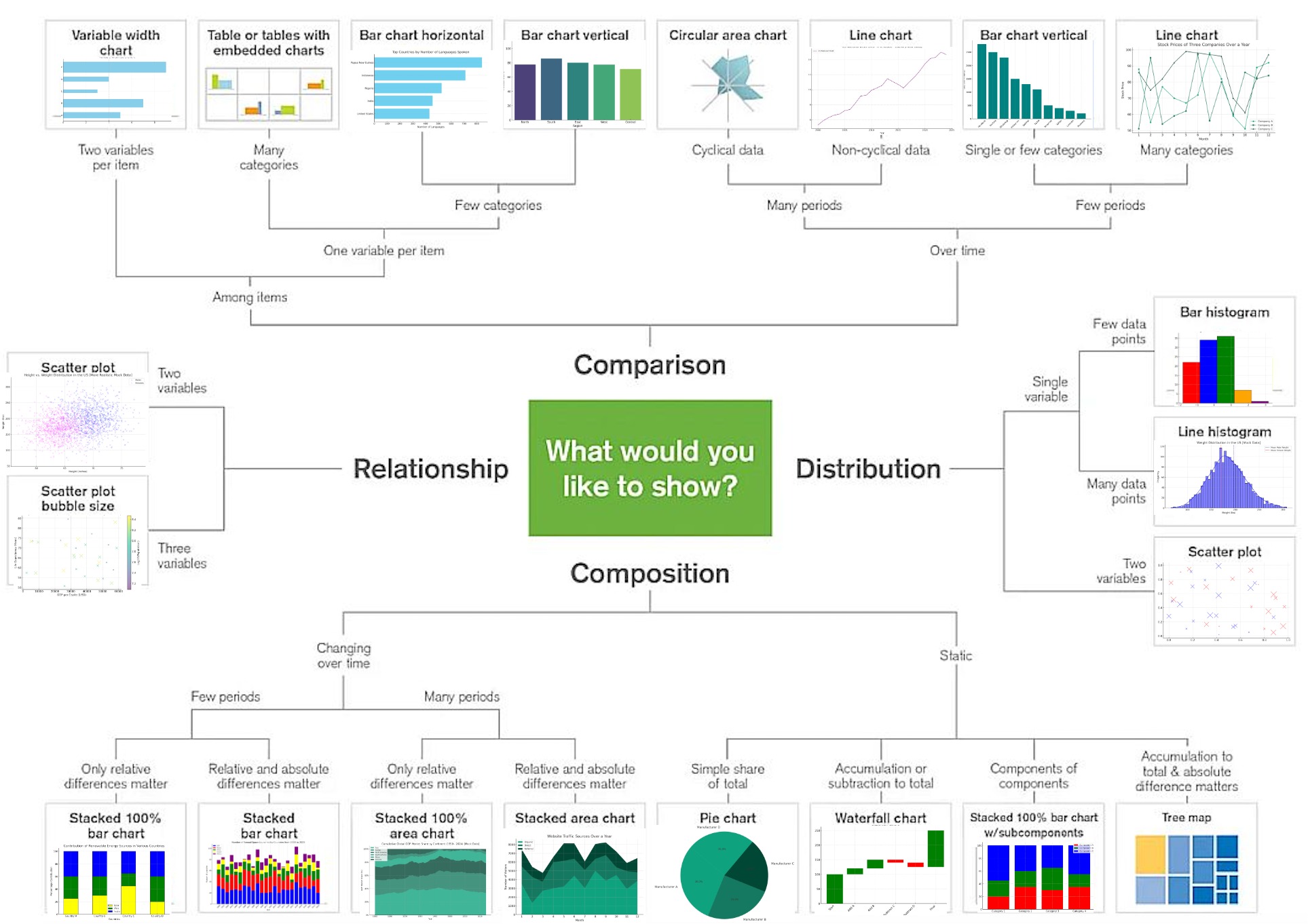}
  \caption{%
  	Chart Cheat Sheet: A visualization cheat sheet to decide which chart is best for a specific type of data and task. All of the charts were created by prompting GPT4 with real-world queries. Notably, and this one of the major findings of our paper, all returned charts matched what is commonly recommended for the query's task and scenario. 
  }
  \label{fig:teaser}
}

\graphicspath{{figs/}{figures/}{pictures/}{images/}{./}} % where to search for the images

\usepackage{tabu}                      % only used for the table example
\usepackage{booktabs}                  % only used for the table example
\usepackage{lipsum}                    % used to generate placeholder text
\usepackage{mwe}                       % used to generate placeholder figures

\usepackage{mathptmx}                  % use matching math font

\begin{document}

%%%%%%%%%%%%%%%%%%%%%%%%%%%%%%%%%%%%%%%%%%%%%%%%%%%%%%%%%%%%%%%%
%%%%%%%%%%%%%%%%%%%%%% START OF THE PAPER %%%%%%%%%%%%%%%%%%%%%%
%%%%%%%%%%%%%%%%%%%%%%%%%%%%%%%%%%%%%%%%%%%%%%%%%%%%%%%%%%%%%%%%

\firstsection{Introduction}

\maketitle

%% \section{Introduction} %for journal use above \firstsection{..} instead
Rapid advances in AI, specifically computer vision and natural language, have led to the development of large language models (LLMs) such as GPT that can generate contextually relevant and high-quality responses in the form of images or text to natural language queries.
LLM models are trained on large amounts of text and image data sourced from a significant portion of the internet which gives them seemingly all-encompassing knowledge. As a result, LLMs have emerged as the de facto source of information for many users on the web.

\begin{figure*}[!t]  
 \centering
 \hspace{11.5mm} 
 \subfloat[\label{subfig:bar1}]{%
    \includegraphics[height=50mm]{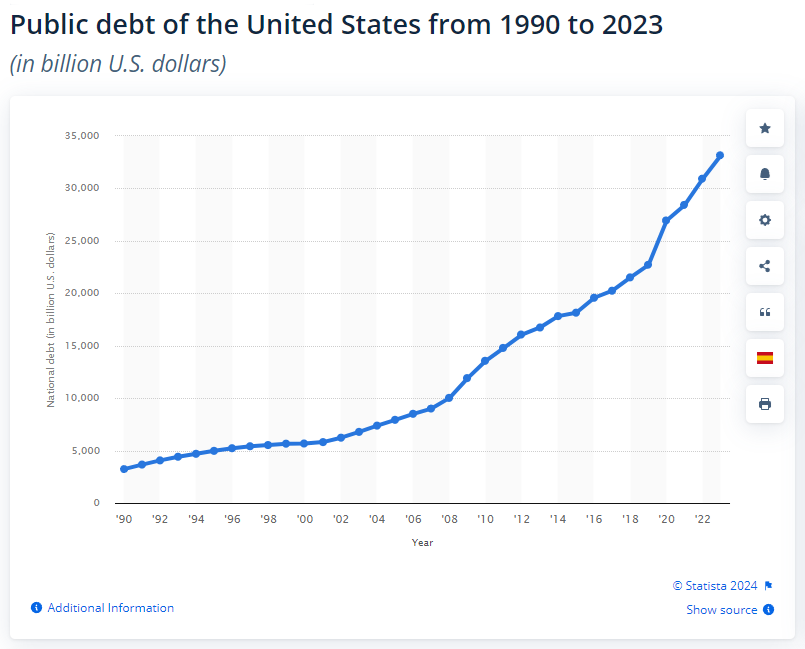}
  }
  \hfill  
  \subfloat[\label{subfig:bar3}]{%
    \includegraphics[height=50mm]{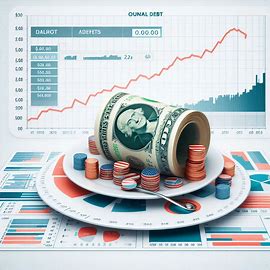}
  }
  \hspace{20mm}
  \vfill
  \hspace{10mm}
  \subfloat[\label{subfig:bar4}]{%
    \includegraphics[height=40mm]{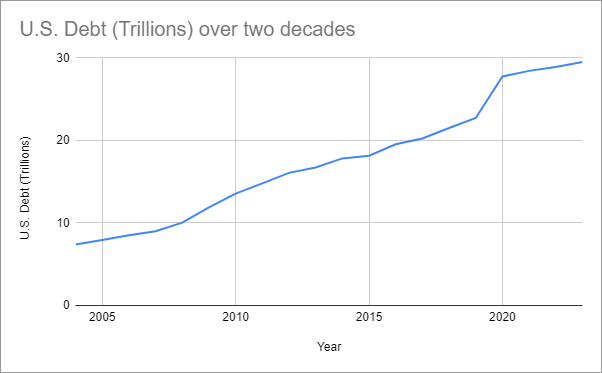}
  }   
  \hfill
  \subfloat[\label{subfig:bar5}]{%
    \includegraphics[height=40mm]{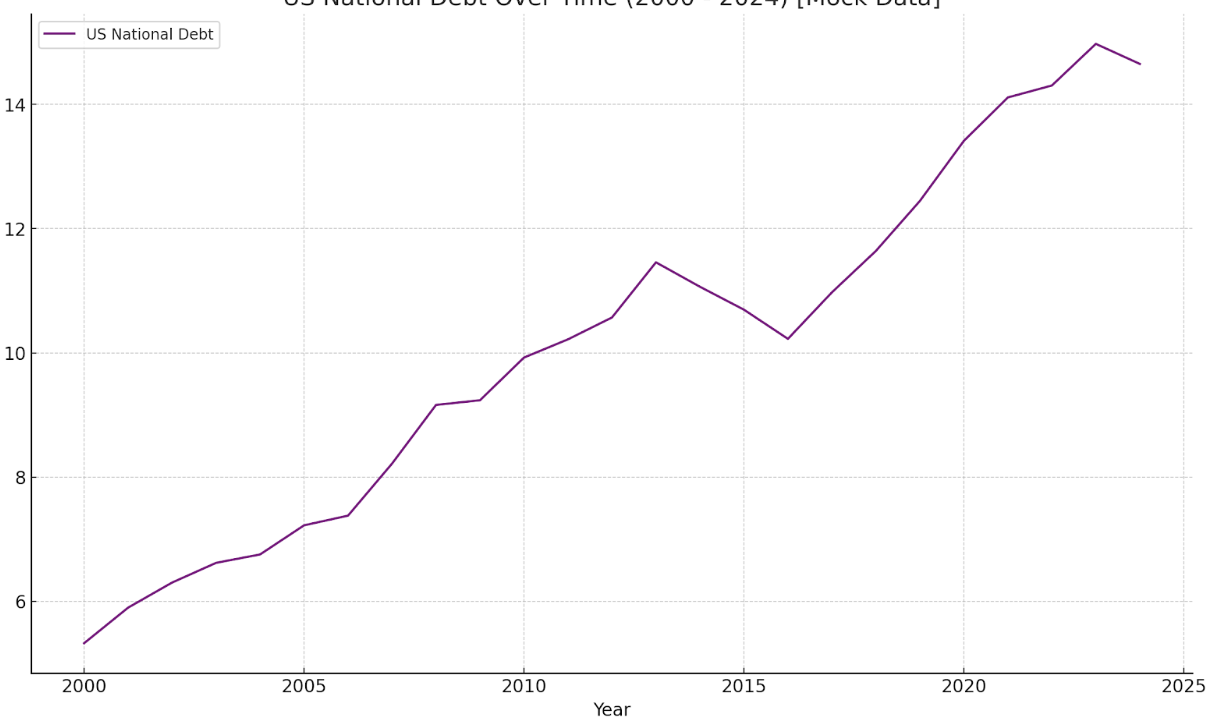}
  }  
  \hspace{10mm}
   \caption{Visualizations showing the U.S. debt over the last two decades. (a) is the ground truth visualization retrieved from Statista while the remaining visualizations are generated with the prompt \textit{"Generate a chart showing the national debt of the U.S. over the last 2 decades"} with (b) DALL-E, (c) GPT-3.5, and (d) GPT-4.}
  \label{fig:initalStudy}  
  \vspace{-5pt}
\end{figure*}

People often have queries about publicly available data on the web, even though they do not possess the data themselves.
For example, they might ask "What is the US debt over the last two decades?" As LLMs are trained on data available on the web, we hypothesize that they have captured publicly available data and can respond to these queries accurately. Additionally, given that visual representations can significantly enhance comprehension \cite{cleveland1984graphical}, we believe that responses accompanied by data visualizations are ideally suited to such queries. 

We expect that when queried for a visual of the data, an LLM would provide us with an appropriate visual of the data as these models have also captured visualization rules \cite{li2024visualization}.
For example, if we prompted an LLM "Generate a chart showing the US debt over the last two decades", we would expect to receive a line chart showing the US debt at each year over the past 20 years. 

In this work, we investigate the ability of generative AI models to respond appropriately to \textit{dataless queries}.
We define dataless queries as text-based queries that can be answered with publicly available data but do not contain the data themselves. 
We expect the model to provide the data, or an approximation of the data, as part of its response. 
Our initial investigation tested the ability of DALL-E 2, GPT-3.5, and GPT-4, to respond to dataless queries. 
We found that GPT-4 appeared to be more adept at answering such queries while responses from DALL-E 2 and GPT-3.5 were inadequate.
This motivated us to conduct a more in-depth investigation of GPT-4s abilities. 
We generated 15 prompts based on popular data visuals on the web, aided by a crowd-sourced survey to gather queries of popular interest.
We then queried GPT-4 using these prompts and evaluated the responses by comparing them to visualization cheat sheets created by visualization experts and human-generated charts on the web. 

In the following. Section 2 provides relevant background. Section 3 details the study we conducted. Section 4 presents a discussion and concludes our paper.

\section{Background}
Generating data-visualizations based on user generated natural language (NL) queries has been a longstanding research topic. 
Early approaches used traditional natural language processing (NLP) techniques such as rule-based and probabilistic grammar-based approaches to decompose NL queries into a set of machine instructions. 
Cox et al. \cite{cox2001multi} created one of the first NL interfaces for visualization using a grammar-based approach to convert NL questions into database queries and visualize the results.
Eviza \cite{setlur2016eviza} employed a probabilistic grammar-based approach and a finite state machine to allow users to interact with a given visualization using NL commands.
NL4DV \cite{narechania2021NL4DV} is a python library that translates NL queries into visualizations. It combines NLP approaches to break down an NL input into the machine instructions required to query a dataset and generate a chart. 

More recently researchers have explored deep-learning based approaches to convert NL queries to visualizations. 
Lui et al \cite{liu2021advisor} proposed ADVISor, a deep-learning based system that uses multiple models to process and transform data and then generate annotated visualizations.
Luo et al \cite{Luo2022Natural} devised a novel approach using a transformer-based model with several novel visualization-aware optimizations to take in a NL query along with an optional chart (e.g., a pie chart or a scatter plot) that serves as a template and generates a visualization. 
RGVisNet \cite{Song2022RGVisNet} is a hybrid system that first looks up a large data visualization query codebase for a query most relevant to the user's data visualization query. 
The retrieved query is then refined by a GNN-based deep-learning model to create a visualization specification.

Recent advances in generative AI, especially LLMs such as OpenAI's GPT \cite{NEURIPS2020_1457c0d6}\cite{openai2024gpt4}, have prompted researchers to investigate their application to the field of data visualization.
Maddigan and Susnjak \cite{Maddigan2023Chat2VIS} showed how ChatGPT and GPT-3 can be leveraged to build an end-to-end solution for converting NL queries to visualizations. 
Their system, Chat2Vis, takes in a NL query and a dataset from the user and behind the scenes engineers prompts to precisely query GPT-3. The queries are constructed such that GPT-3 responds with a python script to visualize the data provided by the user. 
Dibia et al. \cite{dibia-2023-lida} proposed a four-stage approach with LIDA that combines LLMs with image generation to turn datasets and analysis goals into charts and infographics.
Finally Li et al \cite{li2024visualization} evaluated GPT-3.5 for vega-lite specification generation using multiple prompting strategies.

All of these previous approaches required a dataset as input, along with the query to produce a visualization. Conversely, we have studied to what extent LLMs can deliver a visualization with just the query, or prompt, alone. Furthermore, we studied whether the visualizations so produced follow commonly available visualization expert guidelines. In other words, have LLMS learnt both data \textit{and} visualization knowledge?

\begin{figure*}[!ht]
\centering
  \includegraphics[width=0.85\textwidth]{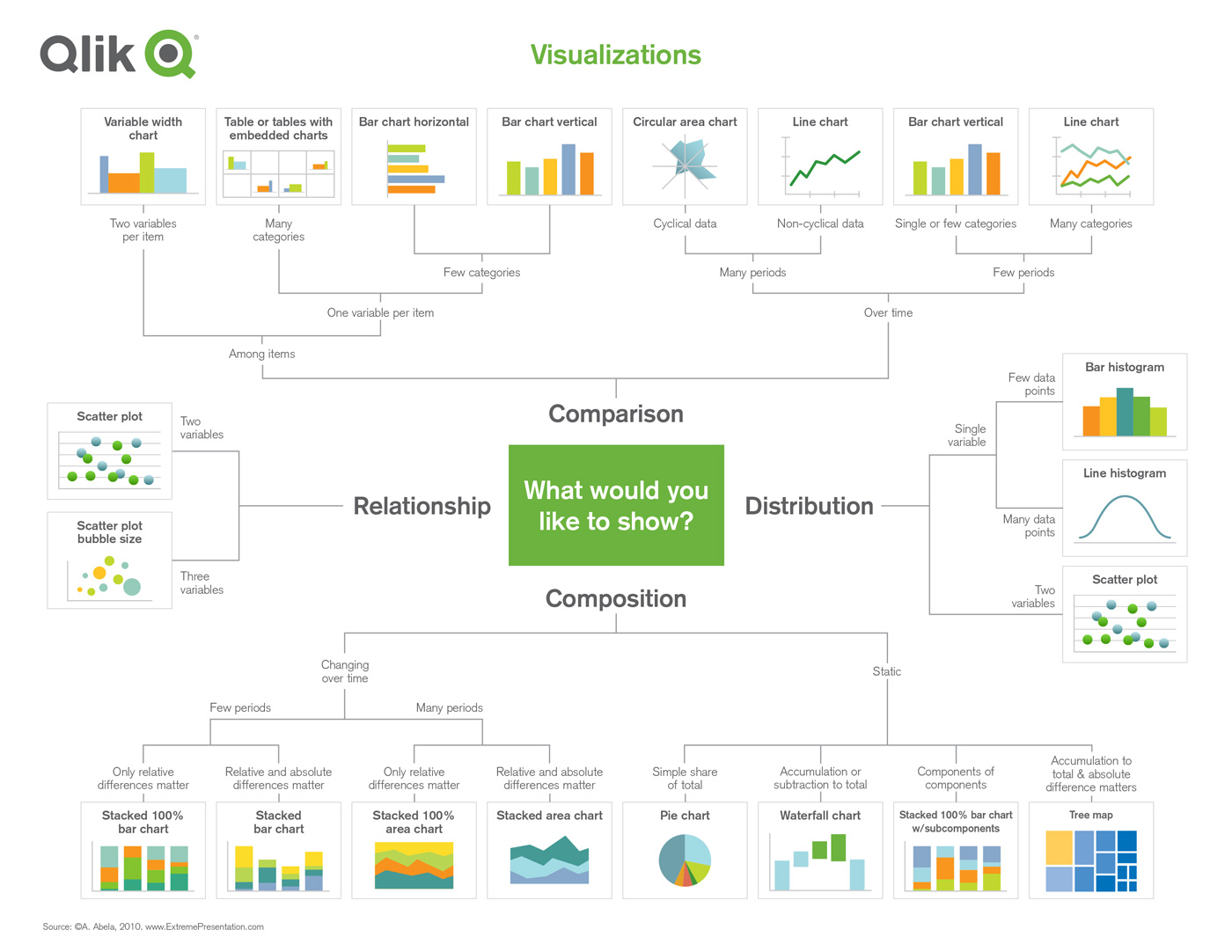}
  \caption{The visualization cheatsheet created by Patrik Lundblad at Qlik to assist visualization designers pick appropriate charts for their data}
    \label{fig:qlikcheatsheet}
\end{figure*}

\section{Study}
Our study aims to investigate the ability of generative AI to produce visualizations when prompted with queries without data. 
This is much like the way a user queries an image search engine for data visualizations e.g. "Generate a plot of the most spoken languages".
In this section, we discuss how we chose to construct \textit{dataless prompts}, how multiple models responded to a small set of dataless prompts, and provide a detailed investigation into how GPT-4 performs on dataless prompts.

\subsection{Dataless Prompts}
We define dataless prompts as text-based queries provided to LLMs that can be answered with publicly available data but do not contain the data themselves. 
An example of a dataless prompt is "What was the US GDP over the past 20 years."
For this prompt, LLMs often provide a text-based response explaining the GDP trends with values for some of the 20 years.
However, providing this prompt to a traditional search engine (e.g. Google or Yahoo) will provide the user with links to articles or images showing the values.
In our work we are interested in the images, specifically the data visualizations. 
To have an LLM produce a visualization, we need to explicitly tell it to provide a chart, for example, "Generate a chart showing the US GDP over the past 20 years."
Given such a prompt, LLMs will respond with an image of a chart or code to generate a chart.
Thus for all dataless prompts in our study, we explicitly ask the models to provide a chart.

\subsection{Initial Investigation}
We began with an initial investigation to test if generative AI modes could produce appropriate responses to dataless prompts.
We tested GPT-3.5, GPT-4 and DALL-E. 
We chose to add DALL-E to the investigation as using it is very similar to how a user would use an image search engine. 
We queried all three models with three prompts and visually inspected the results. 
Each model's response to the query "Generate a chart showing the national debt of the U.S. over last 2 decades" along with a reference ground truth image sourced from Google images is shown in figure \ref{fig:initalStudy}
.
Note that the GPT-3.5 responded with code which we ran on our local machine to generate the chart in figure \ref{fig:initalStudy}c.

In our tests, all three models were able to generate coherent responses. 
GPT-3.5 did not have the capability to respond with an image however responses contained mock data values or pointers to publicly available data and a python code to generate a chart.
GPT-4 could generate charts but it stated that it used mock data. 
However, in both cases, the overall trend of the data was similar to the original data.
DALL-E on the other hand could only generate a chart (shown in figure \ref{fig:initalStudy}b) for one prompt while it generated graphics related to the prompt topic for the others. 
In the case that it did generate the chart, the overall trend did match that of the original data. 
Based on the responses of each model, we concluded that GPT-4 was the only model that could respond with an appropriate image or chart.
GPT-3 was unable to provide a visual result while DALL-E could not provide appropriate visuals.

\subsection{Dataless Prompts with GPT-4}
Having identified GPT-4 as a promising model to respond to dataless prompts, we decided to test it on a wider range of prompts. 

\subsubsection{Procedure}
To test GPT-4's ability to respond to dataless prompts we first generated 15 prompts. 
We generated 7 prompts by asking Google Images for various chart types. 
We then selected 7 random charts and constructed a prompt based on the data used to generate it. 
For example, we selected a pie chart showing Amazon's 2022 revenue breakdown, and subsequently created the prompt "Create a chart to display the revenue breakdown of Amazon in 2022."

To add more variety, we crowd-sourced the remaining 8 prompts. 
To crowd-source the prompts we sent a Google form to graduate students in our university's computer science department asking "If you could ask ChatGPT to visualize public data, what would you ask it to visualize for you? List as many queries as you'd like".
Seven students responded with a total of 22 responses. 
Next, we tested if each response returned a relevant chart on Google images.
We discarded irrelevant responses such as "Make unique recipes using GPT".
This process resulted in us collecting 8 prompts with relevant charts on Google images. 

Once we finalized our set of 15 prompts we fed each prompt to GPT-4 and compared the result to the Google image we collected. 
The result returned by GPT-4 and the Google image result for each of the 15 prompts are shown in the supplementary material. 
For each response, we evaluated if (a) GPT-4 returned a chart with the data we requested, (b) it used an appropriate chart type, and (c) the data matched the data in the chart retrieved with Google Images.

\begin{figure}[!t]  
 \centering
 \subfloat[\label{subfig:bar11}]{%
    \includegraphics[height=60mm]{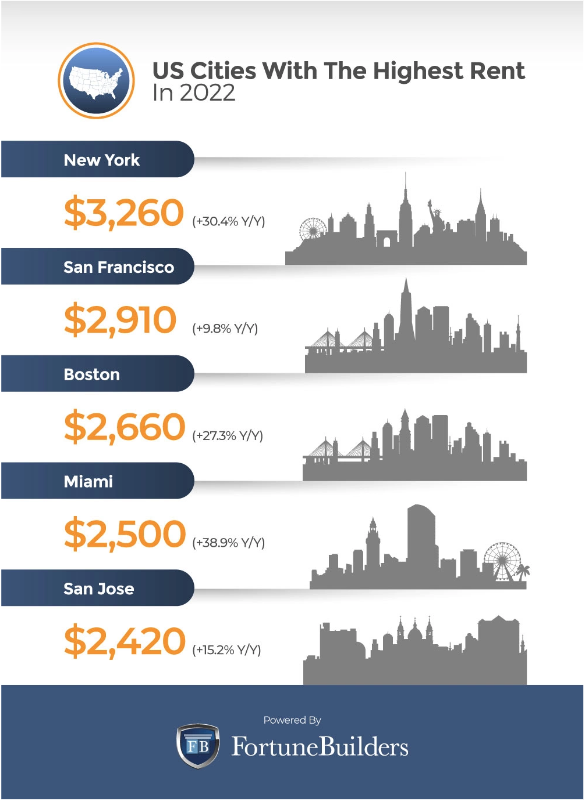}
  }
  \hfill  
  \subfloat[\label{subfig:bar31}]{%
    \includegraphics[height=50mm]{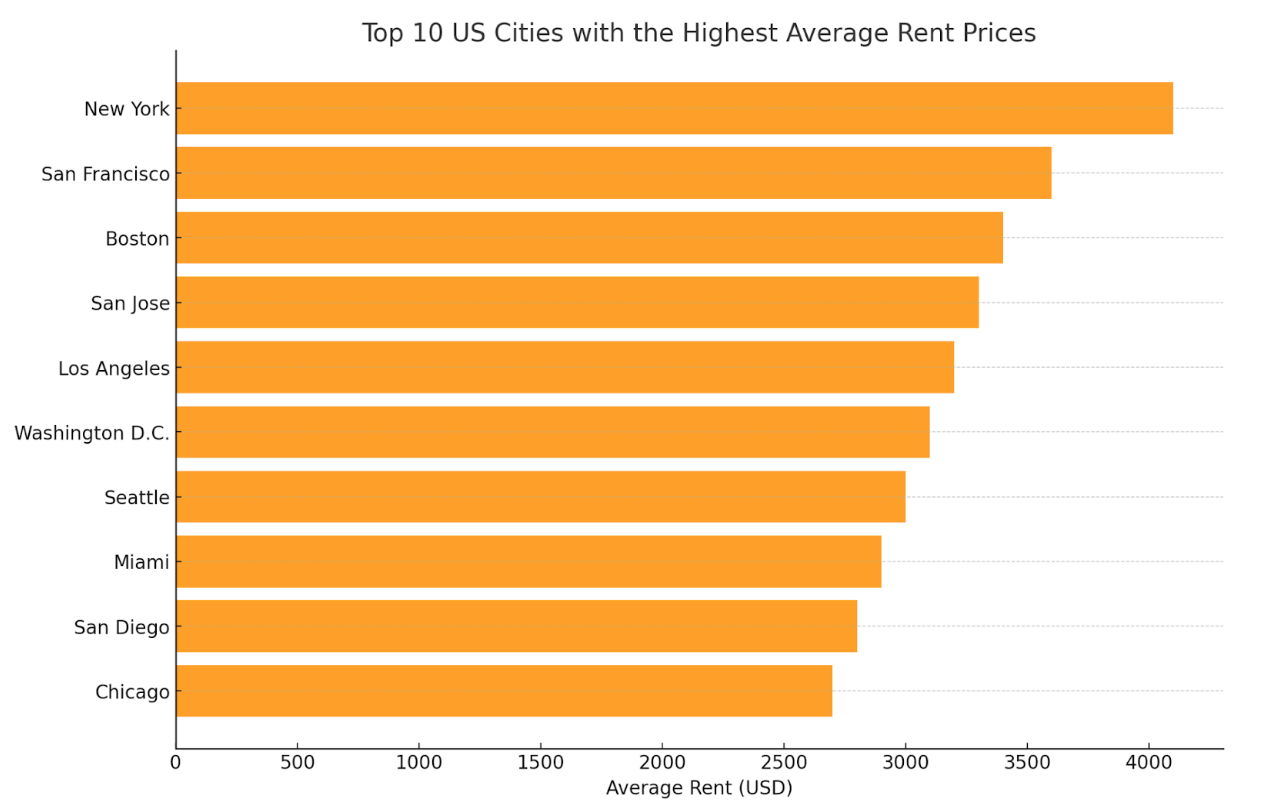}
  }
   \caption{Visualizations retrieved from (a) Google images and (b) GPT-4 showing the U.S. cities with the highest average rent. When compared to the visualization from Google images, we see that GPT-4 was able to generate a similar ordering of cities however it provided different values for mean rents, perhaps it retrieved data from a different year.}
  \label{fig:initalStudy1}  
  \vspace{-5pt}
\end{figure}

\subsubsection{Results}
Overall GPT-4 performed very well, producing coherent responses to all prompts. 
For every prompt, it provided data requested in the prompt. 
We then evaluated whether each returned chart followed commonly accepted visualization design rules, given the specific data and query scenario. For this we used the widely accepted visualization cheat sheet developed by Patrik Lundblad at Qlik \cite{qlikheatsheet} shown in figure \ref{fig:qlikcheatsheet}.
We found that every chart generated was appropriate! This can be gleaned from figure \ref{fig:teaser} which shows the placement of each GPT-4 generated chart into the appropriate slot in Lundblad's cheat sheet.

GPT-4 faltered, however, when it came to reproducing the exact same data shown in the charts retrieved from Google Images. 
When generating line charts GPT-4 could procudce charts with the same general trend, however the values were not identical and the trends were only correct at a larger scale. For example, in figure \ref{fig:initalStudy}d, GPT-4 introduced a dip in the U.S. debt between the years 2013 to 2016 however the original data shown in figure \ref{fig:initalStudy}a shows that the debt was monotonically increasing. 
When it came to generating rankings with bar charts (e.g., top UEFA football clubs, most expensive cities to rent in etc) GPT-4 appeared to be more accurate.
It was able to generate the same overall ranking of items but the values represented by bars were inconsistent. 
See for example figure \ref{fig:initalStudy1}, where a chart from Google images (a) reported the mean rent in New York to be \$3260
while GPT-4 (b) reported it as \$4100.

\section{Discussion and Conclusions}
Our paper makes two contributions. First, we show that GPT-4 can return useful visualizations from dataless queries -- queries that are not accompanied by associated data files, as has been common practice so far. It so absolves mainstream users from having to look for data files first before generating visualizations with GPT-4. While the visualizations so produced are not perfect in every detail, they nevertheless indicate that GPT-4 has obtained a sufficient amount of data knowledge plus the skill to produce a visual rendition of it. 

Second, we find that GPT-4 has been an excellent 'student' of the art and science of data visualization, having learned an impressive amount of visualization knowledge from its 'teachers', the many visualizations found on the web. We put GPT-4 to the test by asking it to fill in a commonly used visualization cheat sheet \cite{qlikheatsheet} where it reached a perfect score (compare figure \ref{fig:teaser} with the cheat sheet published in \cite{qlikheatsheet}). 

During our experimentation, we noticed that GPT-4 seemed to insist on using specific visualizations for specific data queries, namely those shown in the cheat sheet, even when the prompt was moderately altered to create confusion. We value this as an indication that GPT-4's visualization knowledge is fairly well expressed. solid, and ingrained.  

Further, we also found that while GPT3.5 was not as reliable in creating the right type of visualization, it was still good at identifying the sources of data. To that end, a current line of our work is to identify effective mechanisms for LLMs to pull CSV files on its own from the respective online repositories, followed by running the code that generates the appropriate visualizations to provide users with results. Finally, we also believe that there is high promise in Dall-E to eventually be able to produce high quality infographics. We see this as another line of possible important future work.

\bibliographystyle{abbrv-doi-hyperref}

\bibliography{template}

\appendix % You can use the `hideappendix` class option to skip everything after \appendix

\end{document}